\documentclass[runningheads]{llncs}
\usepackage{graphicx}
\usepackage{multirow}
\usepackage{graphicx}
\usepackage{subfigure}
\usepackage{amsmath}
\usepackage{amssymb}
\usepackage[misc]{ifsym}
\usepackage[dvipsnames]{xcolor}
\usepackage[colorlinks,
            linkcolor=green,       
            anchorcolor=green,  
            citecolor=green,        
            ]{hyperref}
\usepackage{float}
\begin{document}
\title{Domain-level Pairwise Semantic Interaction for Aspect-Based Sentiment Classification}
\titlerunning{Domain-level PSI for ABSC}
%
%
%
\author{Zhenxin Wu\and
Jiazheng Gong \and
Kecen Guo\and
Guanye Liang\and\\
Qingliang Che\and Bo Liu\\
}

\authorrunning{Z. Wu et al.}
%
\institute{
\email{2533653096@qq.com, 619790446@qq.com}}

\maketitle              
\begin{abstract}
Aspect-based sentiment classification (ABSC) is a very challenging subtask of sentiment analysis (SA) and suffers badly from the class-imbalance. Existing methods only process sentences independently, without considering the domain-level relationship between sentences, and fail to provide effective solutions to the problem of class-imbalance. From an intuitive point of view, sentences in the same domain often have high-level semantic connections. The interaction of their high-level semantic features can force the model to produce better semantic representations, and find the similarities and nuances between sentences better. Driven by this idea, we propose a plug-and-play Pairwise Semantic Interaction (PSI) module, which takes pairwise sentences as input, and obtains interactive information by learning the semantic vectors of the two sentences. Subsequently, different gates are generated to effectively highlight the key semantic features of each sentence. Finally, the adversarial interaction between the vectors is used to make the semantic representation of two sentences more distinguishable. Experimental results on four ABSC datasets show that, in most cases, PSI is superior to many competitive state-of-the-art baselines and can significantly alleviate the problem of class-imbalance.

\end{abstract}
\section{Introduction}
\begin{figure} 
\centering
\includegraphics[width=0.98\textwidth]{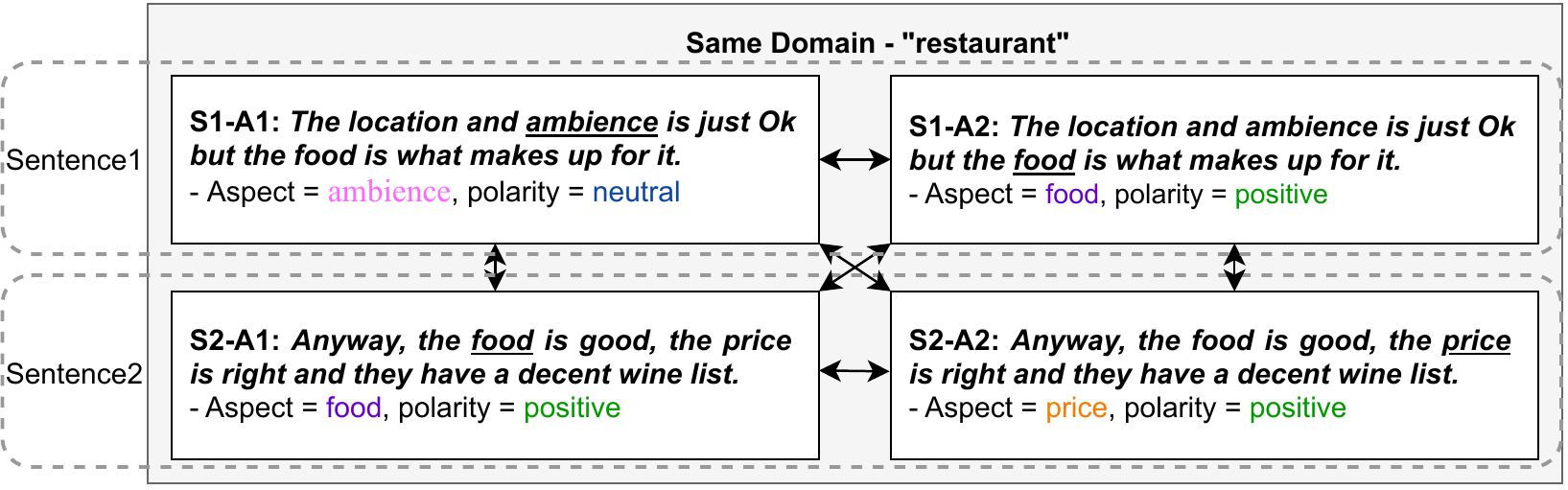}
\caption{The two sentences have different aspects, but they all belong to the same domain “restaurant” and have similar semantic context.}
\label{1}
\end{figure}

Aspect-based sentiment classification (ABSC) is a fine-grained sentiment classification subtask of sentiment analysis~\cite{DBLP:conf/acl/LiHZL10}, which aims to identify the sentiment polarity of each aspect in a sentence (positive , negative or neutral). It is widely used in different domains, such as online comments (e.g., movie and restaurant reviews~\cite{DBLP:conf/semeval/KiritchenkoZCM14}), data mining and e-commerce customer service. For example, sentence 1 in Fig.~\ref{1} shows that the customer enjoys the restaurant’s food but thinks the ambience is just not bad. For this sentence, ABSC needs to recognize that the two aspects “ambience” (A1) and “food” (A2) contained in the sentence are “neutral” and “positive”, respectively. 

In fact, people’s comments often have obvious emotional preferences, which means that they may suffer the problem of class-imbalance. Since there are far more comments with “positive” and “negative” in the same domain than those with “neutral”,  “neutral” comments are always  marginalized and thus misjudged. At present, the commonly used ABSC methods, whether they are traditional methods~\cite{DBLP:conf/semeval/WagnerACBBFT14,DBLP:conf/icml/JinH09} or deep learning models~\cite{DBLP:conf/naacl/DevlinCLT19,DBLP:conf/aaai/WuO21}, none of them has solved the problem of class-imbalance. Moreover, the similarity of semantic contexts between sentences in the same domain has not been fully utilized. In our paper, we define “semantic” as a highly abstract coding vector of sentences extracted by the information extractor, e.g. BERT. If we can make interactive learning of two similar sentences in the same domain, they can learn more domain semantic information from each other and enrich the high-level semantic encoding of sentences. It will also help to find the similarities and nuances between sentences, which can reduce misjudgments due to class-imbalance.

For example, making interactive learning of S1-A1 and S1-A2 in Fig.~\ref{1}, can help to distinguish different sentiment polarities of a sentence which contains different aspects. The interaction between S1-A1 (“neutral”) and S2 (both aspects are “positive”) can make the semantic encoding of “neutral” more discriminative, by comparing it with the strong “positive” sentence S2. At the same time, the interaction between S1-A2, S2-A1, and S2-A2 can also enrich the features of the same sentiment polarity in different aspects or different sentences.

Based on this intuition, we propose a domain-level plug-and-play Pairwise Semantic Interaction (PSI) module for ABSC. For the construction of sentence pairs, it is worth emphasizing that we consider that the sentences in the same dataset belong to the same domain. We do not limit that two sentences must have the same aspect, and encourage richer interactions between sentences (refer to~\ref{3.4} for detailed sentence pair construction strategy). For PSI module, firstly, we extract the semantic vectors of the two sentences by semantic extractors (e.g., BERT~\cite{DBLP:conf/naacl/DevlinCLT19}), respectively. Subsequently, through a gating mechanism, sentences can learn each other’s high-level semantic information adaptively, which enriches the semantic representation of a single sentence. Finally, we additionally use a design similar to the adversarial network~\cite{DBLP:conf/nips/GoodfellowPMXWOCB14} to help promote the model to distinguish the nuances between similar semantic representations. In summary, the contributions of this paper are as follows:
\begin{itemize}
\item We introduce a Pairwise Semantic Interaction (PSI) module for the interaction between sentences, help to find the similarities and nuances of sentences, and can significantly reduce the misjudgments due to class-imbalance. Through the interaction between sentences in the same domain, the sentences get better and more discriminative semantic representations.
\item The PSI module is plug-and-play and can be easily combined with most mainstream semantic extractors such as BERT. 
\item  The experiments on four prestigious ABSC datasets have justified the efficacy of PSI, achieving or approaching SOTA results.
\end{itemize}

\section{Related work}
Existing ABSA researches focus on the use of deep neural networks, such as target dependent LSTM models \cite{DBLP:conf/coling/TangQFL16} and Attention-based LSTM \cite{DBLP:conf/emnlp/WangHZZ16} for aspect-level sentiment classification. In recent years, the pre-trained language model BERT \cite{DBLP:conf/naacl/DevlinCLT19}, which has been very successful in many Natural Language Processing tasks, has been applied in ABSA and achieved significant results such as \cite{DBLP:conf/naacl/SunHQ19,DBLP:conf/naacl/XuLSY19,DBLP:conf/emnlp/LiYZP20}. However, all of the above studies have ignored semantic relationship between sentences in the same domain. Recently, contrastive learning has achieved great success in both Computer Vision (CV) and Natural Language Processing (NLP). Its main purpose is to make the features of the same category closer to each other, while the distance between the features of different categories is farther. In \cite{DBLP:conf/aaai/ZhuangW020}, through an attention interaction, the network can adaptively find delicate clues from two fine-grained images in pairs.

In ABSA, Chen et al. \cite{DBLP:conf/acl/ChenSWLSZZ20} proposed a Cooperative Graph Attention Networks (CoGAN) method for cooperatively learning the aspect-related sentence representation in document level. Tang at al. \cite{DBLP:conf/acl/TangJLZ20} also use the method of transformer combined with graph, in order to allow dependency graph to guide the representation learning of the transformer encoder. However, their model is based on transformer combined with Graph Networks, which has high computational overhead. By contrast, our proposed PSI is a plug-and-play module, which can achieve decent performance with a little extra overhead.

\section{The Proposed Method}

\begin{figure}
\centering
\includegraphics[width=1.0\textwidth]{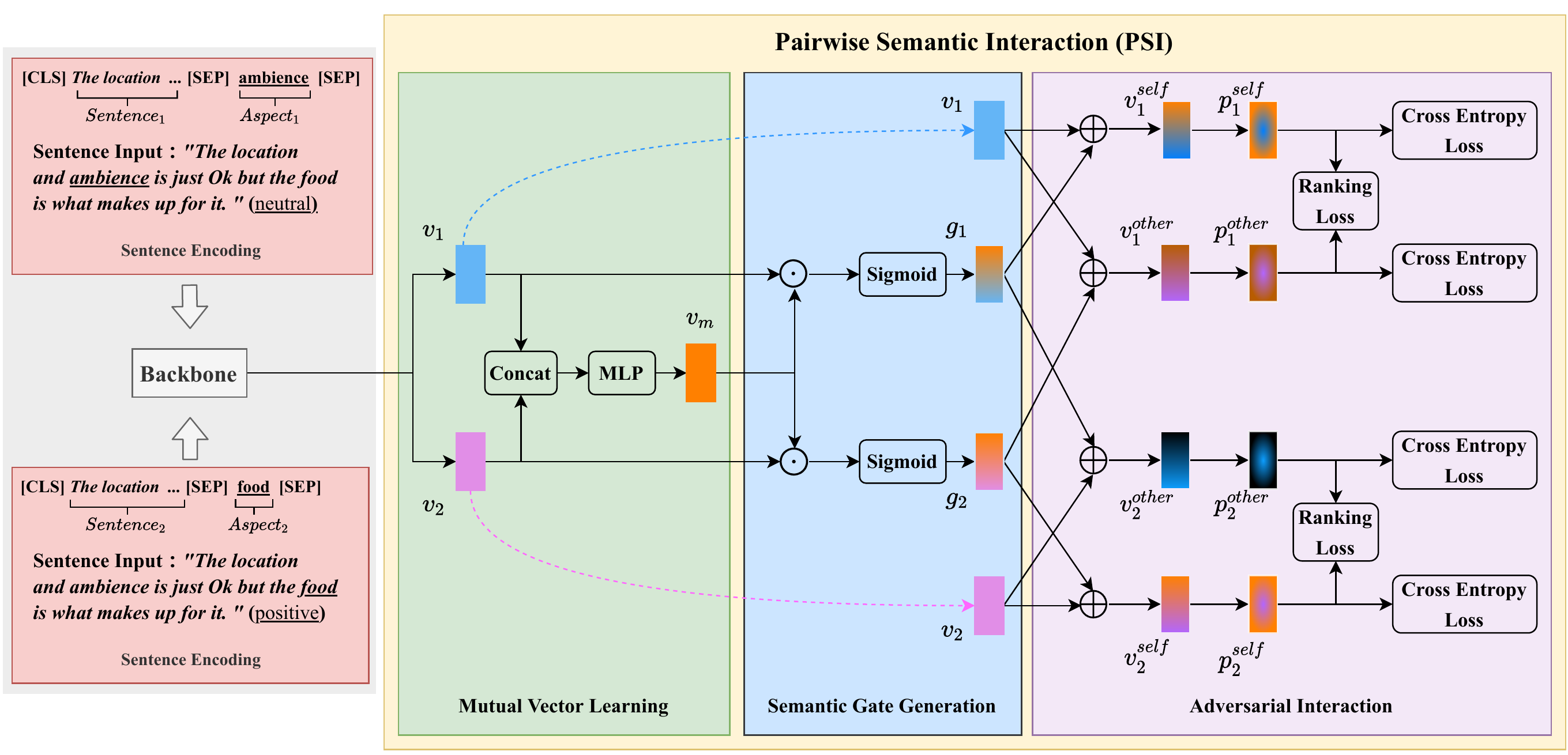}
\caption{The structure of PSI. We simply use S1-A1 and S1-A2 in Fig.~\ref{1} as an example of pairwise sentences ($Sentence_{1}$,$Sentence_{2}$). In this case, PSI can promote the model to distinguish different sentiment polarities of different aspects in the sentence. It is worth emphasizing that the PSI is a plug-and-play module, i.e., PSI can be combined with most mainstream semantic extraction backbones (e.g., BERT) during the training phase, and flexibly unload it for single-input test sentence.} 
\label{2}
\end{figure}

In this section, we will describe our PSI module. PSI compares two similar sentences together to find the common semantic representation and semantic differences between them, rather than studying the semantic representation of a single sentence alone.

The module PSI will take two similar sentences as input and go through three carefully designed sub-modules i.e., mutual vector learning, semantic gate generation, and adversarial interaction. The entire structure of PSI is shown in Fig.~\ref{2}.


\subsection{Mutual Vector Learning}
Before this sub-module, the semantic of these two sentences are extracted by backbone, and two $D$-dimensional semantic vectors i.e., $v_{1}$ and $v_{2}\in \mathbb{R}^{D}$ are generated, respectively. Then, in this sub-module, we learn a mutual vector $v_{m}\in \mathbb{R}^{D}$ from individual $v_{1}$ and $v_{2}$,

\begin{equation}
v_{m}=f_{m}([v_{1},v_{2}]).
\label{001}
\end{equation}

where [ ] is concatenation operation and $f_{m}(\cdot)$ is a mapping function of $[v_{1},v_{2}]$. Specifically, we use the multi-layer perceptron (MLP) as the mapping function. By summarizing the two feature vectors $v_{1}$ and $v_{2}$, the mutual vector $v_{m}$ is produced accordingly, which contains common high-level semantic information and discriminative semantic clues of two sentences.

\subsection{Semantic Gate Generation}
After producing the mutual vector $v_{m}$, we can use $v_{m}$ to activate $v_{1}$ and $v_{2}$. In order to generate more discriminative information for later comparisons, the dot product of $v_{m}$ with two feature vectors $v_{1}$ and $v_{2}$ is carried out according to channels to locate the contrastive information in the two vectors. Then, the gate vectors, i.e., $g_{1}$ and $g_{2}$ are generated by a sigmoid function,

\begin{equation}
g_{i}=sigmoid(v_{m}\odot v_{i}), i\in \left \{ 1,2 \right \}.
\label{002}
\end{equation}

Therefore, $g_{i}$ can be used as an attention vector to highlight the important semantic representations belonging to individual $v_{i}$. For example, the previous example of S1-A1 and S1-A2 in Fig.~\ref{1}, the gates will help to highlight two different key words “ambience” and “food”, respectively. This helps to distinguish different sentiment polarities belonging to different aspects.

\subsection{Adversarial Interaction and Model Training}
When comparing two sentences, humans not only focus on the salient parts of one sentence, but also focus on the salient parts of the other one. Based on this, we introduce an adversarial interaction mechanism through residual attention. As shown in Fig.~\ref{2}, two feature vectors $v_{1}$ and $v_{2}$ and two gate vectors $g_{1}$ and $g_{2}$ are combined in pairs, we could then get four attentive semantic vectors with

\begin{equation}
	\begin{split}
	v_{1}^{self} &= v_{1}+v_{1}\odot g_{1},\\
	v_{2}^{self} &= v_{2}+v_{2}\odot g_{2},\\
	v_{1}^{other} &= v_{1}+v_{1}\odot g_{2},\\
	v_{2}^{other} &= v_{2}+v_{2}\odot g_{1}.
	\end{split}
	\label{003}
\end{equation}

Intuitively, the semantic vector $v_{i}$($i \in \{1,2\}$), is guided by the attention of the gate vector $g_{j}$($j \in \{1,2\}$), strengthens or weakens certain semantic information, and then adds to itself to get the output. While $v_{i}^{self}\in \mathbb{R}^{D}$ reinforces the feature region belonging to its own gate vector, and $v_{i}^{other}\in \mathbb{R}^{D}$ reinforces the feature region belonging to another gate vector.

Then  $v_{i}^{j}$(where $i \in \{1,2\}$, $j \in \{self,other\}$) are feed into the softmax classifier by $p_{i}^{j} = softmax(Wv_{i}^{j}+b)$, where $p_{i}^{j}\in \mathbb{R}^{C}$ represents the score vector of prediction, $C$ indicates the number of polarities, and $\left \{ W,b \right\}$ is the parameter set of the softmax classifier. In order to effectively train the entire PSI module, we define the following loss function as
\begin{equation}
 	J = J_{ce}+\mu J_{rk}.
 \label{005}
\end{equation}

Among them, $J_{ce}$ is the cross-entropy loss, and $J_{rk}$ is the score ranking regularization loss with a coefficient of $\mu $. Specifically we choose the hinge loss function as the score ranking regularization $J_{rk}$,
    \begin{equation}
 	J_{rk} = \sum\nolimits_{i\in \left \{ 1,2 \right \}}\max(0,p_{i}^{other}(y_{i})-p_{i}^{self}(y_{i})+\varepsilon).
 \label{007}
\end{equation}

where $p_{i}^{j}(y_{i})\in \mathbb{R}$ represents the score got in the predicted vector $p_{i}^{j}$, and $y_{i}$ denotes the index of the true polarity of sentence \textit{i}, and $\varepsilon$ is the penalty term. The motivation of this design is that, $v_{i}^{self}$ is activated by its own gate vector. Hence, compared to $v_{i}^{other}$, it should be more discriminative to the corresponding label. That is, the score difference $p_{i}^{self}(y_{i})$-$p_{i}^{other}(y_{i})$ should be larger than a margin $\epsilon$, which means that $p_{i}^{self}(y_{i})$ should be larger than $p_{i}^{other}(y_{i})$ and must keep a distance with $p_{i}^{other}(y_{i})$. At the same time, when cross-entropy loss $J_{ce}$ is optimized, due to having the same label, $p_{i}^{self}(y_{i})$ and $p_{i}^{other}(y_{i})$ will tend to be closer. Therefore, $J_{rk}$ and $J_{ce}$ will be optimized adversarially. As a result, $v_{i}^{other}$ can learn the semantic information shared by the two sentences, and $v_{i}^{self}$ can learn its own unique information which will be more discriminative and reduce the noise of sentence pairs.

\subsection{Sentence Pair Construction}\label{3.4}
Next, we'll provide an explanation on how to construct multiple sentence pairs in a batch for end-to-end training. Specifically, we randomly sample $N_{p}$ polarities in a batch (there are 3 polarities in total, i.e. positive, negative, neutral). For each polarity, we randomly sample $N_{s}$ training sentences. Consequently, there are $N_{p}\times N_{s}$ different sentences in each batch (we set the same sentence to express different aspects, belonging to different sentences). After getting a batch of sentences, we input these sentences into the backbone to generate their respective semantic vectors. For every sentence, we compare its semantic vector with the different sentences in the batch in accordance to Euclidean distance. We do not limit that two different sentences must have the same aspect, and encourage richer interactions between sentences(the following ablation study proves our point). Then, we can construct the inter/intra-pairs in a batch. The inter-pairs are the following sentence pairs which contains two situations. 1) The current sentence and itself (with different aspect and different polarity), e.g., S1-A1 \& S1-A2 in Fig.~\ref{1}; 2) The current sentence and the most similar sentence with different polarities from the current sentence, e.g., S1-A1 \& S2(A1/A2). On the contrary, intra-pairs refer to the following sentence pairs. 1) The current sentence and itself (with different aspect and same polarity), e.g., S2-A1 \& S2-A2; 2) The current sentence and the most similar sentence with the same polarity from the current sentence, e.g., S1-A2 \& S2(A1/A2). This design permits the PSI to learn to distinguish between truly similar and highly overlapping pairs.

\subsection{Model Testing}
Because PSI is a practical plug-and-play module. In the training phase, the backbone and the PSI module can summarize the comparative clues from sentence pairs, and step by step improve the discriminant capacity of backbone representation for sentences. Therefore, in the testing phase, only the backbone model with updated parameters is used, but not the PSI module, so that the generalization ability of the model can be guaranteed without losing the performance of the model. To be specific, in testing phase, we input a sentence into the backbone, extract its semantic vector $X_{*}\in \mathbb{R}^{D}$, and then directly input $X_{*}$ into the softmax classifier. It is worth emphasizing that the softmax classifier are shared between the training phase and the testing phase. The score vector $P_{*}\in \mathbb{R}^{C}$ is applied to label prediction. Thus, our test scheme is the same as a regular backbone, which demonstrates the strong applicability of PSI.

\section{Experiments}
\subsection{Datasets and Metrics}
We have carried out experiments on four datasets to verify the performance of our proposed model, PSI. Restaurant14 is from Semeval-2014 Task 4~\cite{DBLP:conf/semeval/PontikiGPPAM14}, Laptop15 is from Semeval-2015 Task 12~\cite{DBLP:conf/semeval/PontikiGPMA15}, and the other two datasets (Restaurant16 and Laptop16) are from Semeval-2016 Task 5~\cite{DBLP:conf/semeval/PontikiGPAMAAZQ16}. The statistics for these datasets are shown in Table~\ref{table 1}. And we use Accuracy (Acc.) and Macro-F1 (F1) as performance metrics.

\begin{table*}[t]
\centering
\caption{Statistics of the datasets.}\label{tab1}
\begin{tabular}{c|cc|cc|cc|cc}
\hline
\multirow{2}{*}{Polarity} & \multicolumn{2}{c|}{Res14}     & \multicolumn{2}{c|}{Lap15}   & \multicolumn{2}{c|}{Res16}   & \multicolumn{2}{c}{Lap16}    \\
                          & train         & test                & train         & test         & train         & test         & train         & test         \\ \hline
Positive                  & 839           & 222                & 765           & 329          & 749           & 204          & 1084          & 274          \\ \hline
Neutral                   & 500           & 94                      & 106           & 79           & 101           & 44           & 188           & 46           \\ \hline
Negative                  & 2179          & 657                    & 1103          & 541          & 1657          & 611          & 1637          & 481          \\ \hline
Sum                       & \textbf{3518} & \textbf{973}  & \textbf{1974} & \textbf{949} & \textbf{2507} & \textbf{859} & \textbf{2909} & \textbf{801} \\ \hline
\end{tabular}
\label{table 1}
\end{table*}

\subsection{Implementation Details}
Unless stated otherwise, we implement PSI as follows. For each aspect of a sentence, we concat the corresponding aspect at the end of the sentence. Then we adjust the length of each sentence to 85 (the maximum sentence length after the tokenizer tokenizes is 85). If it is not enough, fill it with zero, and then use it as the input of backbone. Firstly, we extract the semantic vector $v_{i}\in\mathbb{R}^{786}$ by BERT. Secondly, for all the datasets, we randomly sample 3 polarities, i.e., $N_{p}=3$. And for each polarity, we randomly sample 4 sentences to form a batch, i.e., $N_{s}=4$. For each sentence, we find its most similar sentence from its own polarity and the rest polarities, according to Euclidean distance between their semantic vectors. As a result, we obtain an intra-pair and an inter-pair for each sentence in the batch. For each pair, we concatenate $v_{1}$ and $v_{2}$ as input to a two-layer MLP, i.e., FC (1572→512), FC (512→786). Consequently, this operation generates the mutual vector $v_{m}\in\mathbb{R}^{786}$.
All of our models are implemented by Pytorch with a single NVIDIA GTX 2080Ti GPU with 11G Memory. For all datasets, the coefficient $\mu$ in Eq.~\ref{005} is $1$, while the margin $\epsilon$ is 0.05 in score ranking regularization. Among them, BERT is optimized by Adam optimizer with $\beta _{1}$= 0.9, and the initial learning rate is 0.0001. For our PSI (backbone is BERT or BERT-Large) method, we use another Adam optimizer for training, and the initial learning rate is 0.00002 with $\beta _{1}$= 0.9. There are a total of 20 training epochs, and if the loss does not decrease for 5 consecutive epochs, it will invoke early-stop. In addition, we set a fixed seed when training the model to ensure the reproducibility of the results.

\subsection{Comparison with SOTA Methods}
To fully evaluate the performance of our method, we apply PSI based on BERT or BERT-Large. We compare it with the state-of-the-art (SOTA) baselines including (1) ABSA models without BERT: TC-LSTM~\cite{DBLP:conf/coling/TangQFL16}, ATAE-LSTM~\cite{DBLP:conf/emnlp/WangHZZ16}, RAM~\cite{DBLP:conf/emnlp/ChenSBY17}, IAN~\cite{DBLP:conf/ijcai/MaLZW17}, Clause-LevelATT~\cite{DBLP:conf/ijcai/WangLLKZSZ18}, LSTM+synATT+TarRep~\cite{DBLP:conf/coling/HeLND18},
 kumaGCN~\cite{DBLP:conf/emnlp/ChenTZ20}, RepWalk~\cite{DBLP:conf/aaai/ZhengZMM20} and IMN~\cite{DBLP:conf/acl/HeLND19}. (2) BERT-based models for ABSA: BERT~\cite{DBLP:conf/naacl/DevlinCLT19}, BERT-QA~\cite{DBLP:conf/naacl/SunHQ19}, AC-MIMLLN~\cite{DBLP:conf/emnlp/LiYZP20} and  CoGAN~\cite{DBLP:conf/acl/ChenSWLSZZ20}. Table~\ref{table2} shows the results of our experiments on four datasets.
 
 From Table~\ref{table2} we can come to the following conclusion. The performance of PSI (Based on BERT or BERT-Large) on Res14, Lap15 and Lap16 is better than those of all baselines. And in Res16, our PSI module approaches SOTA results. The experiments justify that PSI is a very powerful plug-and-play module, showing the effectiveness of our method.
 
\begin{table*}[t]
\centering
\setlength{\tabcolsep}{1mm}
\caption{The comparative results, with data for non-BERT models from \cite{DBLP:conf/acl/ChenSWLSZZ20}, kumaGCN from~\cite{DBLP:conf/emnlp/ChenTZ20}, RepWalk from~\cite{DBLP:conf/aaai/ZhengZMM20}, and IMN from~\cite{DBLP:conf/acl/HeLND19}. The data for BERT-QA from \cite{DBLP:conf/naacl/SunHQ19}, AC-MIMLLN from~\cite{DBLP:conf/emnlp/LiYZP20}, and CoGAN from~\cite{DBLP:conf/acl/ChenSWLSZZ20}. The experimental configuration for standard BERT and PSI is shown in implementation details. “-” denotes no data available yet. The best result of each dataset is bolded, and the second-best result is underlined.}\label{tab2}
\label{table2}
\begin{tabular}{c|cc|cc|cc|cc}
\hline
{\color[HTML]{333333} }                                                                       & \multicolumn{2}{c|}{{\color[HTML]{333333} Res14}}                                            & \multicolumn{2}{c|}{{\color[HTML]{333333} Lap15}}                           & \multicolumn{2}{c|}{{\color[HTML]{333333} Res16}}                             & \multicolumn{2}{c}{{\color[HTML]{333333} Lap16}}                              \\
\multirow{-2}{*}{{\color[HTML]{333333} Models}}                                           & {\color[HTML]{333333} Acc.}           & {\color[HTML]{333333} F1}              & {\color[HTML]{333333} Acc.}          & {\color[HTML]{333333} F1}            & {\color[HTML]{333333} Acc.}           & {\color[HTML]{333333} F1}             & {\color[HTML]{333333} Acc.}           & {\color[HTML]{333333} F1}             \\ \hline
{\color[HTML]{333333} TC-LSTM}                                                                & {\color[HTML]{333333} 0.781}          & {\color[HTML]{333333} 0.675}           & {\color[HTML]{333333} 0.745}         & {\color[HTML]{333333} 0.622}         & {\color[HTML]{333333} 0.813}          & {\color[HTML]{333333} 0.629}          & {\color[HTML]{333333} 0.766}          & {\color[HTML]{333333} 0.578}          \\
{\color[HTML]{333333} ATAE-LSTM}                                                              & {\color[HTML]{333333} 0.772}          & {\color[HTML]{333333} -}              & {\color[HTML]{333333} 0.747}         & {\color[HTML]{333333} 0.637}         & {\color[HTML]{333333} 0.821}          & {\color[HTML]{333333} 0.644}          & {\color[HTML]{333333} 0.781}          & {\color[HTML]{333333} 0.591}          \\
{\color[HTML]{333333} RAM}                                                                    & {\color[HTML]{333333} 0.802}          & {\color[HTML]{333333} 0.708}          & {\color[HTML]{333333} 0.759}         & {\color[HTML]{333333} 0.639}         & {\color[HTML]{333333} 0.839}          & {\color[HTML]{333333} 0.661}          & {\color[HTML]{333333} 0.802}          & {\color[HTML]{333333} 0.627}          \\
{\color[HTML]{333333} IAN}                                                                    & {\color[HTML]{333333} 0.793}          & {\color[HTML]{333333} 0.701}          & {\color[HTML]{333333} 0.753}         & {\color[HTML]{333333} 0.625}         & {\color[HTML]{333333} 0.836}          & {\color[HTML]{333333} 0.652}          & {\color[HTML]{333333} 0.794}          & {\color[HTML]{333333} 0.622}          \\
{\color[HTML]{333333} Clause-Level ATT}                                                       & {\color[HTML]{333333} -}              & {\color[HTML]{333333} -}               & {\color[HTML]{333333} 0.816}         & {\color[HTML]{333333} 0.667}         & {\color[HTML]{333333} 0.841}          & {\color[HTML]{333333} 0.667}          & {\color[HTML]{333333} 0.809}          & {\color[HTML]{333333} 0.634}          \\
{\color[HTML]{333333} \begin{tabular}[c]{@{}c@{}}LSTM+synATT\\ +TarRep\end{tabular}}          & {\color[HTML]{333333} 0.806}          & {\color[HTML]{333333} 0.713}           & {\color[HTML]{333333} 0.822}         & {\color[HTML]{333333} 0.649}         & {\color[HTML]{333333} 0.846}          & {\color[HTML]{333333} 0.675}          & {\color[HTML]{333333} 0.813}          & {\color[HTML]{333333} 0.628}          \\
{\color[HTML]{333333} kumaGCN}                                                                & {\color[HTML]{333333} 0.814}          & {\color[HTML]{333333} 0.736}           & {\color[HTML]{333333} -}             & {\color[HTML]{333333} -}             & {\color[HTML]{333333} 0.894}          & {\color[HTML]{333333} 0.732}          & {\color[HTML]{333333} -}              & {\color[HTML]{333333} -}              \\
{\color[HTML]{333333} RepWalk}                                                                & {\color[HTML]{333333} 0.838}          & {\color[HTML]{333333} 0.769}          & {\color[HTML]{333333} -}             & {\color[HTML]{333333} -}             & {\color[HTML]{333333} 0.896}          & {\color[HTML]{333333} 0.712}          & {\color[HTML]{333333} -}              & {\color[HTML]{333333} -}              \\
{\color[HTML]{333333} IMN}                                                                    & {\color[HTML]{333333} 0.839}          & {\color[HTML]{333333} 0.757}          & {\color[HTML]{333333} 0.831}         & {\color[HTML]{333333} 0.654}         & {\color[HTML]{333333} 0.892}          & {\color[HTML]{333333} 0.71}           & {\color[HTML]{333333} 0.802}          & {\color[HTML]{333333} 0.623}          \\ \hline
{\color[HTML]{333333} BERT}                                                                   & {\color[HTML]{333333} 0.867}          & {\color[HTML]{333333} 0.764}          & {\color[HTML]{333333} 0.818}         & {\color[HTML]{333333} 0.699}         & {\color[HTML]{333333} 0.884}          & {\color[HTML]{333333} 0.755}          & {\color[HTML]{333333} 0.817}          & {\color[HTML]{333333} 0.665}           \\
{\color[HTML]{333333} BERT-QA}                                                                & {\color[HTML]{333333} -}          & {\color[HTML]{333333} -}              & {\color[HTML]{333333} 0.827}         & {\color[HTML]{333333} 0.595}         & {\color[HTML]{333333} 0.896}          & {\color[HTML]{333333} 0.715}          & {\color[HTML]{333333} 0.812}          & {\color[HTML]{333333} 0.596}          \\
{\color[HTML]{333333} \begin{tabular}[c]{@{}c@{}}AC-MIMLLN\end{tabular}}              & {\color[HTML]{333333} 0.893}          & {\color[HTML]{333333} -}              & {\color[HTML]{333333} -}             & {\color[HTML]{333333} -}             & {\color[HTML]{333333} -}              & {\color[HTML]{333333} -}              & {\color[HTML]{333333} -}              & {\color[HTML]{333333} -}              \\ 
{\color[HTML]{333333} \begin{tabular}[c]{@{}c@{}}CoGAN\end{tabular}}              & {\color[HTML]{333333} -}          & {\color[HTML]{333333} -}              & {\color[HTML]{333333} 0.851}             & {\color[HTML]{333333} 0.745}             & {\color[HTML]{333333} \textbf{0.920}}              & {\color[HTML]{333333} \underline{0.816}}              & {\color[HTML]{333333} \underline{0.842}}              & {\color[HTML]{333333} 0.707}              \\\hline
{\color[HTML]{333333} \begin{tabular}[c]{@{}c@{}}PSI (BERT)\end{tabular}} & {\color[HTML]{333333} \underline{0.916}} & {\color[HTML]{333333}  \underline{0.857}}  & {\color[HTML]{333333} \underline{0.860}} & {\color[HTML]{333333} \underline{0.756}} & {\color[HTML]{333333} 0.901} & {\color[HTML]{333333} 0.788} & {\color[HTML]{333333} 0.839} & {\color[HTML]{333333} \underline{0.723}} \\
{\color[HTML]{333333} PSI (BERT-Large)}                                                               & {\color[HTML]{333333} \textbf{0.924}}          & {\color[HTML]{333333} \textbf{0.863}}           & {\color[HTML]{333333} \textbf{0.868}}         & {\color[HTML]{333333} \textbf{0.760}}         & {\color[HTML]{333333} \underline{0.913}}          & {\color[HTML]{333333} \textbf{0.828}}          & {\color[HTML]{333333} \textbf{0.87}}          & {\color[HTML]{333333} \textbf{0.737}}          \\
\hline
\end{tabular}
\end{table*}

\subsection{Alleviating the Problem of Class-Imbalance}
As shown in Table~\ref{table 1}, the mainstream datasets have the problem of class-imbalance. For example, in Res14, the “negative” comments is significantly more than the data of other polarities (“negative” accounts for 62\%), while the “neutral” comments is far lower for other polarities (“neutral” accounted for 14\%). In order to illustrate the advantages of the our method, we compared PSI (Based on BERT) with the standard BERT model on the Res14 dataset, for each polarity (positive, negative, and neutral). Table~\ref{table_3} shows the comparison results of the accuracy of different sentiment polarities.


\begin{table*}[t]
\centering
\setlength{\tabcolsep}{3mm}
\caption{Comparison of accuracy(\%) of different polarities between PSI and BERT.}
\label{table_3}
\begin{tabular}{c|c|c|c|c}
\hline
Model & Negative(\%) & Neutral(\%) & Positive(\%) & Overall(\%) \\ \hline
BERT  & 92.5     & 47.9    & 86.0     & 86.7    \\
PSI (BERT)   & \textbf{97.6}(+5.1)     & \textbf{60.6}(+12.7)    & \textbf{86.9}(+0.9)     & \textbf{91.6}(+4.9)    \\ \hline
\end{tabular}
\end{table*}
It can be seen that the “positive” and “neutral” accuracy of our model (PSI) is better than that of BERT model. Specifically, PSI significantly improves the performance of “neutral” classification. And the overall accuracy of our model was greatly improved compared with BERT model. It indicates that the sentence pair interaction learning in PSI module can make the semantics between sentences complement each other and effectively learn the nuances between sentences. This can make the semantic representation of different polarities of sentences more distinguishable, and can effectively alleviate the problem of class-imbalance.

\subsection{Sample Extraction Strategies}
Our proposed method encourages richer interactions between sentences and do not limit that two different sentences must have the same aspect. In order to study the impact of different sample extraction methods (sentiment polarity and aspect) on ABSC, we conducted the following ablation experiments. We use BERT as semantic vector extractor to evaluate different sample extraction methods on Res14. 

As mentioned above, our proposed sample extraction method is that Intra/inter pairs are constructed from the same/different sentiment polarities without limiting the range of aspect (\textbf{Interacting Polarity, $I\_P$}). And there are three other sample extraction methods, including \textbf{1) Interacting Aspect ($I\_A$)}. Intra/inter pairs are constructed from the same/different aspects without limiting the range of sentiment polarity. \textbf{2) Interacting Polarity and Limiting Aspect ($I\_P$ \& $L\_A$)}. Intra/inter pairs are constructed from the same/different polarities by limiting the same aspect. \textbf{3) Interacting Aspect and Limiting Polarity ($I\_A$ \& $L\_P$).} Intra/inter pairs are constructed from the same/different aspects by limiting the same polarity. 

From Table~\ref{tab6}, in all the four datasets, the results of the other three ablation experiments are worse than $I\_P$ (Ours). For $I\_P$ \& $L\_A$ and $I\_A$ \& $L\_P$, results demonstrate that we should not limit aspects (in order to better distinguish different aspects of a sentence) and should also allow different polarities to interact in pairs (in order to better distinguish different sentiment polarities). For $I\_A$, due to class-imbalance, if we do not limit the range of sentiment polarity (by constructing the inter-pair of different sentiment polarities), there will be a lot of interactions between sentences belonging to the same majority class (“negative”), while the interaction between different sentiment polarities will be insufficient. Finally, for $I\_P$, we explicitly construct inter-pairs (belonging to different polarities) in each batch to ensure that the number of interactions between different polarities is sufficient. In this way, the 
nuances between different sentiment polarities can be better learned by the model. Therefore, we choose $I\_P$ as the sample extraction method for ABSC.
\begin{table*}[t]
\centering
\caption{Different Sample Extraction Method.}
\setlength{\tabcolsep}{1mm}
\label{tab6}
\begin{tabular}{c|cc|cc|cc|cc}
\hline
\multirow{2}{*}{{\color[HTML]{333333} Methods}} & \multicolumn{2}{c|}{{\color[HTML]{333333} Res14}}                             & \multicolumn{2}{c|}{{\color[HTML]{333333} Lap15}}                             & \multicolumn{2}{c|}{{\color[HTML]{333333} Res16}}                             & \multicolumn{2}{c}{{\color[HTML]{333333} Lap16}}                              \\
 &{\color[HTML]{333333} Acc.}            & {\color[HTML]{333333} F1}             & {\color[HTML]{333333} Acc.}            & {\color[HTML]{333333} F1}             & {\color[HTML]{333333} Acc.}            & {\color[HTML]{333333} F1}             & {\color[HTML]{333333} Acc.}            & {\color[HTML]{333333} F1}             \\ \hline
{\color[HTML]{333333} I\_P (Ours)}                    & {\color[HTML]{333333} \textbf{0.916}} & {\color[HTML]{333333} \textbf{0.857}} & {\color[HTML]{333333} \textbf{0.860}} & {\color[HTML]{333333} \textbf{0.756}} & {\color[HTML]{333333} \textbf{0.901}} & {\color[HTML]{333333} \textbf{0.788}} & {\color[HTML]{333333} \textbf{0.839}} & {\color[HTML]{333333} \textbf{0.723}} \\
{\color[HTML]{333333} I\_A}                         & {\color[HTML]{333333} 0.914}          & {\color[HTML]{333333} 0.854}          & {\color[HTML]{333333} 0.834}          & {\color[HTML]{333333} 0.699}          & {\color[HTML]{333333} 0.896}          & {\color[HTML]{333333} 0.753}          & {\color[HTML]{333333} 0.830}          & {\color[HTML]{333333} 0.680}          \\
{\color[HTML]{333333} I\_P \& L\_A}                 & {\color[HTML]{333333} 0.909}          & {\color[HTML]{333333} 0.852}          & {\color[HTML]{333333} 0.840}          & {\color[HTML]{333333} 0.699}          & {\color[HTML]{333333} 0.893}          & {\color[HTML]{333333} 0.787}          & {\color[HTML]{333333} 0.819}          & {\color[HTML]{333333} 0.656}          \\
{\color[HTML]{333333} I\_A \& L\_P}                 & {\color[HTML]{333333} 0.895}          & {\color[HTML]{333333} 0.826}          & {\color[HTML]{333333} 0.836}          & {\color[HTML]{333333} 0.689}          & {\color[HTML]{333333} 0.873}          & {\color[HTML]{333333} 0.738}          & {\color[HTML]{333333} 0.820}          & {\color[HTML]{333333} 0.641}          \\ \hline
\end{tabular}
\end{table*}

\subsection{Ablation Study}
In addition, in order to investigate the properties of our proposed PSI, we use Bert as a semantic vector extractor to evaluate its key design on Res14. For fairness, when exploring different strategies for one design, we use the other designs as the base strategies described in the proposed approach and implementation details.

\textbf{Mutual Vector Generation.} To demonstrate the essentiality of $x_{m}$  in Eq.~\ref{001}, we investigate different operations to generate it. \textbf{1) Individual Operation}. The key of $x_{m}$ is to learn mutual information from both sentences in the pair. For comparison, we introduce a baseline without it. Specifically, we replace mutual learning in Eq.~\ref{001} by individual learning $\tilde{x} _{i} = f_{m}(x_{i})$, and use $\tilde{x}_{i}$  to generate the gate vector $g_{i} = sigmoid(\tilde{x}_{i})$ where $i\in \left \{ 1,2 \right \}$ . \textbf{2) Elementwise Operations}. We perform a number of widely-used elementwise operations to generate $x_{m}$, including Subtract Square: $x_{m} = {(x_{1} - x_{2})}^{2}$, Sum: $x_{m} = (x_{1} + x_{2})$, and Product $x_{m} = (x_{1} \ast x_{2})$. \textbf{3) Interactive MLP}. It is the mapping function described in the proposed approach. As shown in Table~\ref{tab5}, the Individual operation (i.e., the setting without $x_{m}$) performs worst. Hence, it is necessary to learn mutual context by $x_{m}$. Based on experimental results, we choose the simple but effective Interactive MLP to generate the mutual vector in our experiments.

\begin{table}[]
\centering
\setlength{\tabcolsep}{3.5mm}
\caption{Different operations of mutual vector.}
\label{tab5}
\begin{tabular}{l|cc}
\hline
{\color[HTML]{333333} Mutual Vector}   & {\color[HTML]{333333} ACC}            & {\color[HTML]{333333} F1}             \\ \hline
{\color[HTML]{333333} Individual}      & {\color[HTML]{333333} 0.893}          & {\color[HTML]{333333} 0.824}          \\
{\color[HTML]{333333} Sum}             & {\color[HTML]{333333} 0.909}          & {\color[HTML]{333333} 0.850}          \\
{\color[HTML]{333333} Product}         & {\color[HTML]{333333} 0.904}          & {\color[HTML]{333333} 0.847}          \\
{\color[HTML]{333333} Subtract Square} & {\color[HTML]{333333} 0.910}          & {\color[HTML]{333333} 0.857}          \\
{\color[HTML]{333333} Interactive MLP}             & {\color[HTML]{333333} \textbf{0.916}} & {\color[HTML]{333333} \textbf{0.857}} \\ \hline
\end{tabular}
\end{table}

\textbf{The Influence of the Number of Interactions.} On the basis of sample extraction method $I\_P$, We further studied the influence of the number of sampling polarity $N_{p}$ and the number of corresponding sentences $N_{s}$ in each batch. The results are shown in Table~\ref{table 4}. It can be seen that PSI is more sensitive to polarity-number than sentence-number. This is because more polarities often lead to richer diversity of sentence pairs. If 3 polarities are selected in a batch, sentence pairs must include the sentence belonging to minority category (neutral). So the sentences belonging to minority polarity (neutral) can learn semantic information from other majority polarity (positive or negative), thereby improving the generalization ability of the model. Therefore, we choose the best setting in our experiments, i.e., $N_{p}=3$ and $N_{s}=4$.
\begin{table}
\centering
\setlength{\tabcolsep}{2mm}
\caption{The influence of the num of polarity \& sentence in each sampling.}\label{tab4}
\begin{tabular}{c|cccccc}
\hline
($N_{p}$, $N_{s}$) & (2,3) & (2,4) & (2,5) & (3,3) & (3,4)          & (3,5) \\ \hline
Acc.                  & 0.892 & 0.894 & 0.895 & 0.914 & \textbf{0.916} & 0.897 \\
F1                        & 0.819 & 0.833 & 0.825 & 0.848 & \textbf{0.857} & 0.832 \\ \hline
\end{tabular}
\label{table 4}
\end{table}

\textbf{The Influence of Sentence Similarity.} Furthermore, We also examine the influence of sentence similarity for intra-pairs and inter-pairs on the Res14 dataset based on $I\_P$. For 12($N_{p}\times N_{s}$) different sentences in each batch, we construct intra/inter-pairs for each sentence, respectively. Therefore, there are 24 sentence pair in a batch. The selection strategies including  1)\textbf{ Random.} We randomly sampled 24 sentence pairs(intra/inter pairs) in a batch; 2) \textbf{Sentence-Distance.} We sampled 24 sentence pairs in accordance to the Euclidean distance between sentences in intra/inter pairs (i.e., Similar (S), Dissimilar (D)). Thus, 8 different Class-Polarity-Sentence settings were generated in Table~\ref{table 3}.

\begin{table}[]
\centering
\setlength{\tabcolsep}{2mm}
\caption{The influence of sentence similarity.}\label{tab3}
\begin{tabular}{c|cc|cc}
\hline
Pair Construction                        & Intra & Inter & Acc.           & F1             \\ \hline
Random                                   & -     & -     & 0.905          & 0.842          \\ \hline
\multirow{8}{*}{Sentence-Distance} & -     & D     & 0.9            & 0.846          \\
                                         & -     & S     & 0.91           & 0.847          \\
                                         & D     & -     & 0.882          & 0.809          \\
                                         & S     & -     & 0.909          & 0.853          \\
                                         & D     & D     & 0.897          & 0.826          \\
                                         & S     & D     & 0.91           & 0.849          \\
                                         & D     & S     & 0.892          & 0.827          \\
                                         & S     & S     & \textbf{0.916} & \textbf{0.857} \\ \hline
\end{tabular}
\label{table 3}
\end{table}

From Table~\ref{table 3}, we can see that most of results of Sentence-Distance are better than Random, which shows that when constructing sentence pairs, the polarity of the sentence and the similarity between them must be considered. Secondly, in the setting of Sentence-Distance, Whether inter or intra, the results of (S) is higher than (D) in most cases. This proves that similar sentence pairs can increase the training difficulty of the model, which can effectively help model to identify subtle semantic differences.

\textbf{The Necessity of Ranking Regularization.} In this section, we conduct experiments on the necessity of ranking regularization in the sub-module (Adversarial Interaction) of PSI. We compared the standard BERT, PSI without Ranking Regularization $J_{rk}$ (based on BERT) and PSI (based on BERT) in five datasets. It can be seen from the Table~\ref{table1} that compared with the standard BERT, the performance of the BERT with PSI(with or without $J_{rk}$) is significantly improved, which proves the effectiveness of the PSI structure. Furthermore, the PSI with $J_{rk}$ is better than PSI without $J_{rk}$. This demonstrates that, without ranking regularization, the effectiveness of adversarial interaction will be greatly reduced. When without ranking regularization, the model is only optimized by cross-entropy loss $J_{ce}$, which will make the semantic vectors ($v_{i}^{self}$/$v_{i}^{other}$) activated by different gates lose their distinction. On the contrary, adding ranking regularization can keep semantic vectors activated by different gates at a distance. This allows the semantic vectors activated by the other's gate to have the common semantic representation of the two sentences ($v_{i}^{other}$), and the semantic vectors activated by the own gate to have unique semantic representation ($v_{i}^{self}$), which is more discriminative. In summary, the PSI with ranking regularization helps model to find the similarities and differences between sentence pairs better, and reduce the noise of sentence pairs, which is effective and necessary.

\begin{table*}[h]
\centering
\small
\caption{The impact of ranking regularization.}
\label{table1}
\begin{tabular}{c|cc|cc|cc|cc}
\hline
\multirow{2}{*}{Method} & \multicolumn{2}{c|}{Res14}   & \multicolumn{2}{c|}{Lap15}    & \multicolumn{2}{c|}{Res16}      & \multicolumn{2}{c}{Lap16}       \\
~ & Acc.           & F1              & Acc.          & F1            & Acc.           & F1             & Acc.           & F1             \\ \hline
BERT                    & 0.867          & 0.764                  & 0.818         & 0.699         & 0.884          & 0.755          & 0.817          & 0.665           \\
PSI without $J_{rk}$        & 0.907          & 0.851                   & 0.837         & 0.719         & 0.888          & 0.718          & 0.833          & 0.714          \\ 
PSI with $J_{rk}$                    & \textbf{0.916} & \textbf{0.857}  & \textbf{0.860} & \textbf{0.756} & \textbf{0.901} & \textbf{0.788} & \textbf{0.839} & \textbf{0.723} \\ \hline
\end{tabular}

\end{table*}

\subsection{Visualization Analysis}
\begin{figure}[htbp]
\centering
\subfigure[\label{051}BERT.]{
\begin{minipage}{0.5\textwidth}
\centering
\includegraphics[width=6cm]{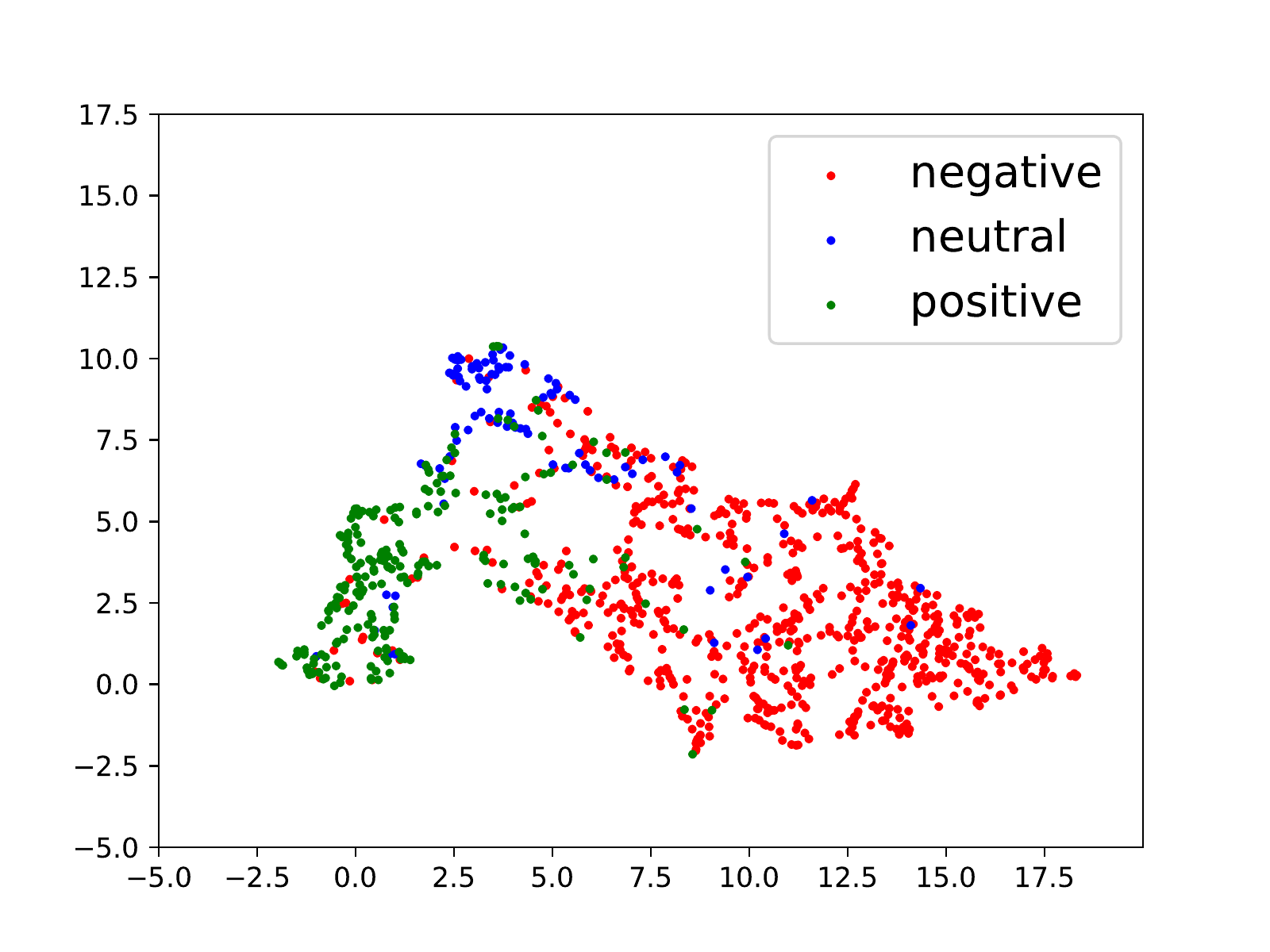}
\end{minipage}%
}%
\subfigure[\label{052}PSI (BERT).]{
\begin{minipage}{0.5\textwidth}
\centering
\includegraphics[width=6cm]{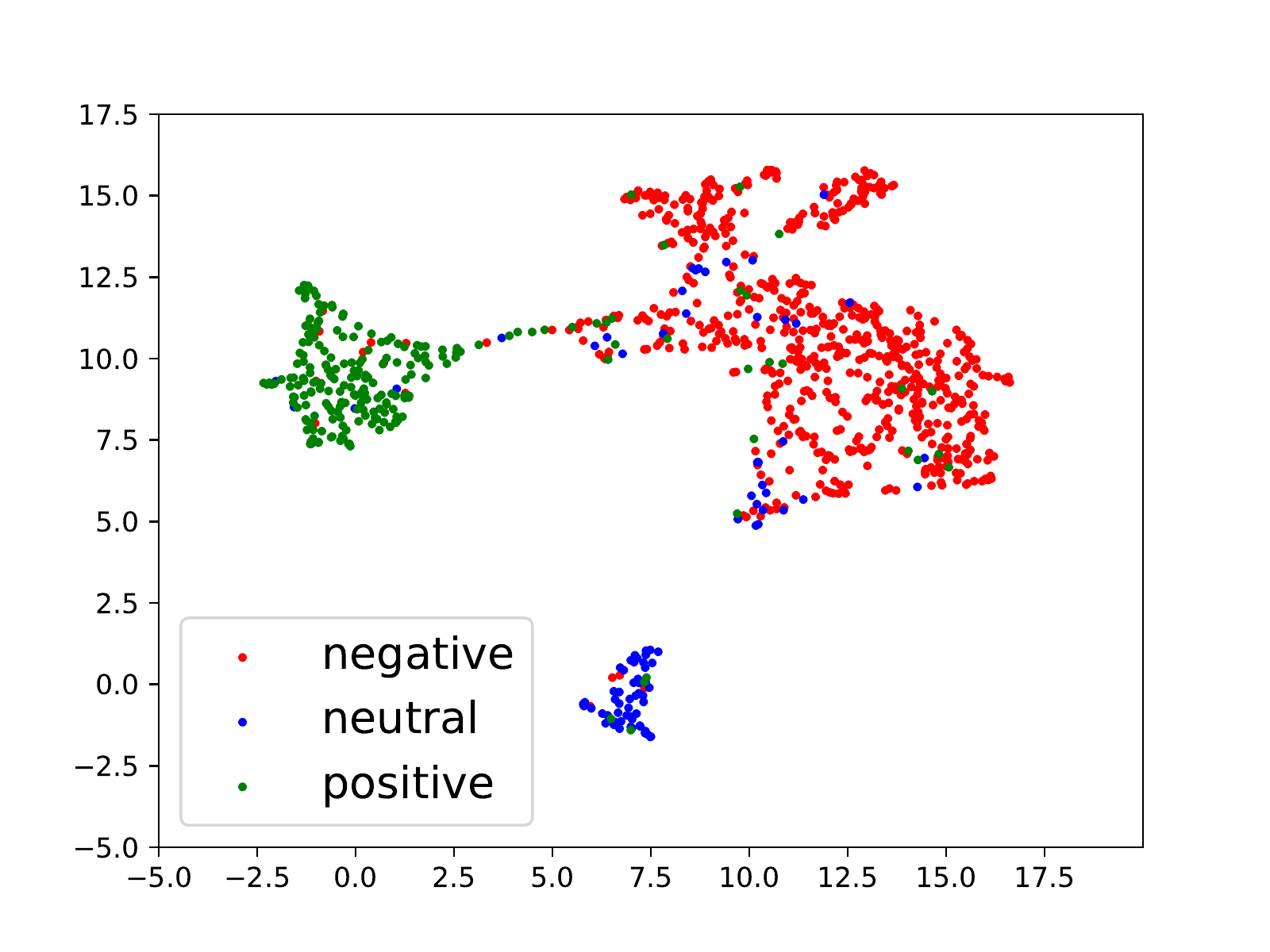}
\end{minipage}%
}%
\caption{Discriminability using UMAP to visualize polarity separability and compactness.}
\label{final}
\end{figure}

In order to understand the discriminability of our method, we use UMAP~\cite{DBLP:journals/corr/abs-1802-03426} to visualize the polarity separability and compactness in the semantic features extracted from a standard BERT and the PSI (based on BERT) in Res14. In Fig.~\ref{final}, it is evident that when using our PSI module, the clusters are farther apart and more compact, leading to a more clear distinction of various clusters representing different polarities. This also proves that adding PSI module can promote the model to learn better semantic representation of sentences and make them more distinguishable.

\section{Conclusion}
In this paper, we proposed a domain-level Pairwise Semantic Interaction (PSI) for ABSC. Through the interactions between sentences, PSI can effectively enrich the semantic encoding of sentences and produce better semantic representations. Meanwhile, PSI is plug-and-play module and can further help the model distinguish the nuances between similar sentences and effectively alleviate the problem of class-imbalance. Finally, the empirical results on four prestigious ABSC datasets justified the power of PSI that has achieved SOTA performance in most cases. In future work, we will consider integrating some advanced attention mechanisms into this method.
\\

\bibliographystyle{splncs04}
\bibliography{mybibliography}
\end{document}